\documentclass[runningheads]{llncs}
\usepackage{graphicx}
\usepackage{booktabs} 
\usepackage{hyperref}

\usepackage[frozencache,cachedir=.]{minted}
\usepackage{wrapfig}
\usepackage[disable]{todonotes}
\usepackage{amsmath}
\usepackage{amsfonts}
\usepackage{scalerel}
\usepackage{mathtools} 
\usepackage{subcaption}
\usepackage{paralist}
\usepackage{multirow}
\usepackage{fancyvrb}
\usepackage{tikz}

\DeclareMathOperator*{\agg}{\scalerel*{\mathsf{A}}{\sum}} 
\newcommand{\eg}{\textit{e.g.}}
\newcommand{\ie}{\textit{i.e.}}
\newcommand{\tomas}[1]{{\color{orange}{Tomas: #1}}}

\newcommand{\citep}[1]{\cite{#1}}
\newcommand{\citet}[1]{\cite{#1}}

\definecolor{OliveGreen}{cmyk}{0.64,0,0.95,0.40}

\begin{document}
\title{Explaining Classifiers Trained on Raw Hierarchical Multiple-Instance Data}
%
%
\author{Tomáš Pevný\inst{1,2} \and
Viliam Lisý\inst{1,2} \and
Branislav Bošanský\inst{1,2} \and
Petr Somol\inst{1} \and
Michal Pěchouček\inst{1,2}}

\authorrunning{T. Pevný et al.}

\institute{Avast Software, Prague, Czech Republic \and
  Artificial Intelligence Center\\
  Department of Computer Science, FEE\\
  Czech Technical University in Prague\\
  \email{<name>.<surname>@avast.com}}
\maketitle              

\begin{abstract}
Learning from raw data input, thus limiting the need for feature engineering, is a component of many successful applications of machine learning methods in various domains. 
While many problems naturally translate into a vector representation directly usable in standard classifiers, a number of data sources have the natural form of structured data interchange formats (e.g., security logs in JSON/XML format).
Existing methods, such as in Hierarchical Multiple Instance Learning (HMIL), allow learning from such data in their raw form.
However, the explanation of the classifiers trained on raw structured data remains largely unexplored.
By treating these models as sub-set selections problems, we demonstrate how interpretable explanations, with favourable properties, can be generated using computationally efficient algorithms. 
We compare to an explanation technique adopted from graph neural networks showing an order of magnitude speed-up and higher-quality explanations.
\end{abstract}

\section{Introduction}\label{sec:intro}
\todo[inline]{Stress importance of scalability. [BB] I have added a sentence.}
\todo[inline]{VL: Currently heavily focused on HMIL rather than explainability. Move to RW or remove?}


\noindent 
One of the reasons behind the success of modern machine learning (ML) models is the ability to process raw input data such as images, text, or sound without hand-designed features. Processing raw input data not only removes the
human work needed to engineer features, 
it also avoids possible human bias in deciding which parts of input data are relevant.
This is confirmed by a significant increase in ML performance 
whenever representation learning became embedded in model learning, \eg, in 
image recognition~\citep{he2015delving}, game play~\citep{silver2017mastering}, or translation~\citep{wu2016google}.

In many domains with successful applications of ML models, the raw input data are in the form of matrices (images) or streams of numbers (text), or they can be easily transformed into such a format. 
Other domains, however, rely on structured heterogeneous and variable-sized data where the transformation to a fixed-sized matrix representation is not straightforward, and, 
more importantly, it would induce loss of information implicitly present in the structure.
Consider, \eg,  e-commerce relational databases storing data about customers, where each record contains a different number of purchases consisting of a varying number of items per purchase.
In cyber-security, data representing executable files are stored and analyzed for the purpose of malware detection. Such data include, \eg, the list of dynamically imported libraries, and for each library, the list of imported functions \citep{ember}.
The executable itself is composed of a variable number of sections and data directories that hold an arbitrary number of data items of various lengths. Other domains ripe for adopting structured featureless ML include engineering design, health care administration, software engineering, various fields of chemistry, etc. 
Data in these domains can be naturally structured as variable-length lists of items or key-value pairs, which recursively consist of other such lists or eventually basic data types, such as numbers, strings, or boolean values. 
The abundance of such data is mirrored in the wide adoption of standard data interchange formats like JSON or XML, directly designed to store this type of data.

There are recent models capable of training over raw structured data including sets (\eg, Deep Sets~\citep{zaheer2017deep}), graph-based structures (cf. Graph Neural Networks~\citep{scarselli2008graph}) and
for arbitrarily nested sets (cf. Hierarchical Multiple Instance Learning (HMIL)~\citep{pevny2016discriminative}).
The idea behind these models is to follow the structure of the data and recursively create their problem-specific fixed-size embeddings as part of model training. 
While these models' performance has been shown to meet or exceed the performance of the best unstructured models working with hand-crafted features \citep{duvenaud2015convolutional,qi2017pointnet,pevny2020nested}, there are still gaps in their deeper understanding and explanation of their decisions.

Accordingly, our main focus is on providing an \textbf{explanation of classifiers trained on raw hierarchical structured data}. 
As an explanation, we seek \textbf{the minimal subset of input} (i.e., a  sub-tree) that is still \textbf{classified to the same class as the complete input sample}. We call the latter condition the \textbf{explanation consistency}.
Consider a sample in the JSON format in Figure~\ref{fig:json}
depicting a device record from a network scan.
This sample should be classified as an audio device by a device type classification model.
The explanation in such a case can determine that, \eg, {\scriptsize \mintinline{json}{{"upnp": [{"manufacturer": "Sonos, Inc."}]}}} is the sufficient part of input
causing this classification.

\begin{figure}[t]
\begin{minted}[fontsize=\scriptsize]{json}
{"mac": "00:0e:58:fc:40:60",  
 "ip": "192.168.254.24",
 "services": [{"port": 1400, "protocol": "tcp"},
	       ...
              {"port": 5353, "protocol": "udp"}],
 "upnp": 
   [{"model_name": "Sonos Play:3",
     "model_description": "Sonos Play:3",
     "manufacturer": "Sonos, Inc.",
     "device_type": "urn:schemas-upnp-org:device:ZonePlayer:1",
     "services": ["urn:upnp-org:serviceId:AlarmClock",
                  "urn:upnp-org:serviceId:MusicServices",
                        ...
	              "urn:tencent-com:serviceId:QPlay"]},
    {"model_name": "Sonos Play:3",
     "model_description": "Sonos Play:3 Media Server",
     "manufacturer": "Sonos, Inc.",
     "device_type": "urn:schemas-upnp-org:device:MediaServer:1",
     "services": ["urn:upnp-org:serviceId:ContentDirectory",
                  "urn:upnp-org:serviceId:ConnectionManager"]},
    ...
\end{minted}
\vspace{-7mm}
\caption{A part of an example data sample.}\label{fig:json}
\end{figure}

An explanation in the form of a minimal subset of raw input features has several of the desirable properties identified in prior work \cite{melis2018towards}. It is \textbf{explicit} (\ie, immediate and understandable), since it uses the raw observations included in the data without any transformations. It is \textbf{faithful} (\ie, the presence of a feature in explanation is indicative of its true importance), since all features included in the explanation are necessary for correct classification. Note that this is not true if the subset of features is not minimal. This form of explanation is also advocated in \citet{carter2019made}, and small explanations are the very definition of the comprehensibility in \citet{Andrews1995}. On the other hand, an explanation of this form does not provide the complete picture about the classification since there may be features, the addition of which would change the classification \citep{mothilal2020explaining}. Finding these would be an interesting complement to the explanations we seek, but it is beyond the scope of this paper.

Having explanations of verdicts by classifiers trained from raw data is essential for several reasons. First, they help debugging and maintenance as well as detecting a possible classifier bias.
Poorly trained classifiers can mistakenly focus on irrelevant signals present in training data 
(\eg, network device classes separable by a timestamp of the scan).
Second, such explanations increase trust in the classifier's recommendation to highly accountable professionals such as cyber-security or healthcare experts who need to understand, verify, and document all supporting information for audit. 
Finally, regulations such as GDPR grant to any person, whose data is used for automated decision making, the right to ``meaningful information about the logic involved'' in the respective decision~\citep{regulation2016regulation}(Art. 15(h)).
Note that the size of raw structured input data can be extensive (\eg, a single JSON describing malware behavior in a sandbox can easily have several hundreds of megabytes); hence scalable explainability is necessary to guarantee the usability of these models in practice. 


The contributions of this paper are: 
(1) Relating the problem of explanation in tree-structured data to subset selection problem and the corresponding algorithms;
(2) a set of algorithms realizing various trade-offs between the speed and the precision (size) of the explanation;
(3) a synthetic dataset with clearly defined ground truth and realistic data distributions suitable for the evaluation of explanation techniques; 
(4) quantitative comparison of the algorithms showing that exploitation of the hierarchical structure of the data substantially increases the speed of explanations with only a minor penalty in explanation size;
(5) showcasing the explanations produced by the algorithms on real-world problems of device identification and malware analysis.

\section{Related Work}

Explainability of ML models has received considerable attention in recent years. The methods considered range from creating simplified versions of the full models \cite{ribeiro2016should}, identifying parts of the input that highly influence classification \cite{zeiler2014visualizing}, through training additional models classifying neural network activation patterns \cite{kim2018interpretability}, to arguing that machine learning models should be explainable by design \cite{rudin2019stop}. Nice summaries of the recent developments in the field are available, e.g., in \cite{ying2019gnnexplainer,lipton2018mythos,montavon2018methods}. Here we focus on explaining structured data by identifying the key subsets of the input samples. 
Carter et al. \cite{carter2019made} also argue for explainability in the form of minimal subsets of features. However, they consider only tabular and text data and only a single method for subset search, which is quadratic in the number of features and hence much more expensive than the methods proposed in this paper. They do not deal with structured data, which present new opportunities and challenges, such as providing explanations in complex raw data, which would traditionally be encoded in a lossy way to fixed feature vectors by domain experts, or hierarchical pruning of the samples. 
There are very few works focusing on structured data, and we are not aware of any work explicitly designed for tree-structured data. 
The nearest related work is 
\emph{GNN Explainer}~\citep{ying2019gnnexplainer,huang2020graphlime}, designed to explain decisions of Graph Neural Network models. 
As such, it is capable of providing explanations on general graphs as well as tree-structured data. However, GNNs need to be extended with a complex hierarchy of types for subsets of nodes. Furthermore, we show that by focusing on the special case of tree-structured data, we achieve significantly better explanation quality with an order of magnitude lower computational cost while still addressing a broad range of industrial and business applications. 
\section{Background}

We use JSON files as data samples; however, our approach is domain-independent and applicable to any structured data exchange format.
A JSON file is a hierarchical structure (see Fig.~\ref{fig:json}) with three types of nodes: (1) \textbf{Dictionary nodes} consist of a set of key-value pairs, where the key is a string and the value can be any other node. (2) \textbf{List nodes} contain an arbitrary number of other nodes with the same structure. (3) \textbf{Atomic nodes} hold a single value, e.g., a string, a number, or a boolean value. The files have a hierarchical structure signifying that data in one sub-tree are more closely related than data in separate sub-trees. For example, in Fig.~\ref{fig:json}, the data under the top-level ``services'' key comes from a port scan, while the data under ``upnp'' is from Universal Plug and Play.

A JSON file can be considered a sequence of characters, but such representation ignores the information contained in the structure and assumes the importance of the ordering of subsections or items in lists, which are often irrelevant.
Therefore, we follow an alternative paradigm to capture the variable size of inputs and see them as sets of substructures. 

\subsection{Hierarchical Multiple Instance Learning}\label{sec:hmil}
Hierarchical Multiple Instance Learning (HMIL) \citep{pevny2016discriminative} is a generalization of the set learning problems~\citep{pevny2017using,zaheer2017deep} that is suitable for hierarchical structured data. The set and sequence learning paradigms are combined and applied recursively. 

\begin{figure}[t]
{\scriptsize
\begin{Verbatim}[commandchars=\\\{\}]
\textbf{[Dict]} (present 57906 times)
  \textcolor{OliveGreen}{mac}: String (10000+ unique out of 57906)
  \textcolor{OliveGreen}{ip}: String (9498 unique out of 56316)
  \textcolor{OliveGreen}{services}: \textbf{[List]} (present 49793 times)
    \textbf{[Dict]} (present 281555 times)
      \textcolor{OliveGreen}{port}: Int64 (7854 unique out of 281555)
      \textcolor{OliveGreen}{protocol}: String (2 unique out of 281555)
  \textcolor{OliveGreen}{upnp}: \textbf{[List]} (present 22881 times)
    \textbf{[Dict]} (present 35616 times)
      \textcolor{OliveGreen}{device_type}: String (96 unique out of 35616)
      \textcolor{OliveGreen}{manufacturer}: String (183 unique out of 35261)
      \textcolor{OliveGreen}{model_description}: String (706 unique out of 22382)
      \textcolor{OliveGreen}{model_name}: String (2913 unique out of 35359)
      \textcolor{OliveGreen}{services}: \textbf{[List]} (present 32712 times)
        String (141 unique out of 78924)
  \textcolor{OliveGreen}{dhcp}: \textbf{[List]} (present 13505 times)
    \textbf{[Dict]} (present 13505 times)
                            ...
\end{Verbatim}
}
\vspace{-5mm}
\caption{A part of the schema for the Device ID dataset.}\label{fig:schema}
\end{figure}

For any set of hierarchical structured data samples, it is possible to create a \textbf{schema} of the dataset -- a global view on the types and structures of data appearing in individual sub-structures of the data samples (see Figure~\ref{fig:schema}).
In the case of JSON files, the schema is defined recursively. For a top-level dictionary node, it includes all keys that appeared on the top level in any of the data samples. Each key is associated with a separate sub-schema constructed recursively from the union of the values below the key in all data samples. For a top-level list node, we assume for the clarity of exposition that all items in the list follow the same schema\footnote{This assumption can be relaxed by creating an intermediate dictionary node with keys for each type of item.}. Then all items in the top-level lists from all data samples are considered to be separate data samples, and the schema of an item in the list is created recursively from this new dataset. 
A schema of an atomic node stores its type. 
The schemas for all the used datasets are in Appendix~\ref{app:schemas}.

When evaluating an HMIL model, the children of each node of a hierarchical data sample are recursively embedded into an automatically optimized fixed-size vector representation and then aggregated to a single embedding of the whole sub-sample rooted in that node. The embeddings and the aggregations are differentiable parametric functions associated with a node of the problem's schema. All nodes in the data sample that correspond to the same node in the schema, \eg, all \texttt{port-protocol} pairs in the list under the \texttt{services} key in Fig.~\ref{fig:json}, are embedded and aggregated using the same instances of the functions. 
\todo{For HMIL propagation purposes (supporting this part of the story), it can be useful to also put a HMIL NN that corresponds to a JSON schema into the appendix.}

The HMIL model creates a fixed-size embedding for any variable-size hierarchical data sample, be it a single path to an atomic node or a tree with millions of leaves in various sub-trees.
For a specific classification or regression problem, this embedding function is complemented by a few more arbitrary neural network layers suitable for the specific purpose. HMIL models have been successfully used in practical applications~\citep{liu2016hierarchical,pevny2016discriminative,janisch2019cost} and are universal function approximators, even when using only the mean as the aggregation function~\citep{pevny2019approximation}.

For the purpose of this paper, it is sufficient to understand that a trained HMIL model can process an arbitrary input that follows the schema. Moreover, any part of the sample defined in the schema \textbf{may be missing}. If processing a sample requires an embedding of a missing sub-tree, a suitable (automatically learned) imputation associated with the corresponding schema node is used.


\subsection{Graph Neural Networks}

Graph neural networks (GNNs) are also, in principle, capable of representing and solving classification problems on JSON files. GNNs store a vector of values in each node of the graph and iteratively repeat the following steps: (1) compute a ``message'' for each pair of related (typically neighboring) vertices; (2) for each node, aggregate the messages related to this node using an aggregation function; and (3) update the vector in each node based on the aggregate. Similar to HMIL, all three steps may be performed by a learnable parametric function.

In order to process a set of JSONs using GNNs, we would still need to create some form of a schema of the problem, which is the key concept of HMIL missing in GNNs.  Nodes of the same type would share the parameters of their message computation functions. A JSON file would be translated into a graph. A node representing a JSON list would have edges leading to the nodes representing each list element. A node representing a dictionary would have an edge for each used key, and each of these edges would have a separate message computation function given by the schema based on the used key. Atomic nodes would include a fixed representation of the data, just as in HMIL. The standard inference methods for GNNs would repetitively propagate the information to all nodes in each iteration, rather than performing a single pass from the leaves to the root as in HMIL models. However, since GNNs allow using a different message computation function in each iteration, the single-pass behavior can be emulated using empty messages. The resulting model would formally be a GNN, but it would heavily rely on the concept of schema coming from HMIL, and it would just emulate the computation performed by HMIL in a less intuitive way.

\section{Method}
For clarity of presentation, we assume a binary classification problem over hierarchical structured data and explain why a sample belongs to the positive class. We assume that an HMIL model $h$ embeds the sample to $R^m$ and then a parametric function $f: \mathbb{R}^m \rightarrow [0,1]$ returns the confidence that it belongs to the positive class. The model predicts the positive class for sample $s$ if $f(h(s)) \geq 0.5$.

The reference implementation for all the algorithms and the experimentation scripts, hyper-parameters, and datasets used will be available on github and are included in the supplementary materials.

Finding an explanation for a sample consists of two high-level phases. First, all sub-trees in a sample are heuristically ranked to reflect their importance for the final classification. Second, subtrees of a sample are searched through and evaluated by the model to find the minimal explanation. The following two sub-sections include alternative definitions of these two phases with different performance and applicability trade-offs.

\subsection{Sub-tree Ranking Methods}\label{sec:ranking}

\textbf{Model gradient ranking.~~}
Important input features usually produce a high absolute value of gradients in the model. This is the basis of explanations based on saliency maps~\citep{simonyan2013deep}, and it was used as a baseline in~\citet{ying2019gnnexplainer}. In the case of an HMIL model, we can compute the gradient of the model with respect to the embedding of any of the data sample sub-trees.
For a sample $s$ and its sub-tree $c$, let the gradient of the classifier's output w.r.t. the sample be
$g = \frac{\partial f(h(s))}{\partial h(c)}  \in \mathbb{R}^m$
for some dimension of sub-tree $c$ embedding $m$. Then the model gradient value of the sub-tree is the sum of gradient coordinates $grad(c) = \left|\sum_{i=1}^m g_i \right|.$
The value indicates how much the output of the model on the sample is influenced by a change in the embedding of the sub-tree.

\textbf{GNN explainer mask ranking.~~}
GNN explainer~\citep{ying2019gnnexplainer} is a method for explaining decisions of graph neural networks. Since HMIL can be seen as a special case of GNN, we use GNN explainer as one of the baselines in our experiments. GNN explainer optimizes a real-valued mask controlling how much information is passed along each edge. The mask contains a value between zero and one for each edge, which is used as a weight in the aggregation step of the GNN. If GNN explainer is asked to provide an explanation of a classification decision on a particular node, the mask is optimized using stochastic gradient descent in order to maximize the probability of correct classification on that node. Furthermore, the mask's sparsity is promoted by adding minimization of the entropy and sum of elements as regularization to the objective.
After the mask is optimized, GNN explainer suggests using the subgraph induced by the $k$ edges corresponding to the $k$ highest values in the mask as the explanation. Note that if the $k$ is fixed in advance, the explanation \textbf{may not satisfy the explanation consistency}, \ie, the explanation may be classified into a different class than the original sample. Such an explanation would often not be faithful because it could include additional features, which are unnecessary. 
The GNN explainer mask values we use for ranking the sub-trees in our algorithms are the mask's values for the edges connecting the sub-trees to their predecessor. 

\textbf{Banzhaf values.~~}
%
%
%
Banzhaf values are a notion originating in cooperative game theory, similar to Shapley values, expressing how much a certain player contributes to various coalitions on average \citep{banzhaf1964weighted}. In feature selection, these values have been used to assess the importance of individual features for a classifier~\citep{somol2011fast,sun2012feature}. 
To approximate the Banzhaf values, we use a sampling algorithm similar to those analyzed in \citet{Bachrach2010}. We estimate the value for all tree nodes at once. For each node in the explained data sample, we store two values: the average output of the classifier in coalitions that include the node and the average value of coalitions that do not include the node. 
We repetitively generate a uniform random subset of nodes in the sample by independently deciding to include or omit each node in the tree. We compute the output of the classifier for the subset. Afterward, we update one of the values stored in each node, depending on whether the node is present in the evaluated subset. After running a larger number of iterations, the Banzhaf value approximation for each node is the difference between the two values stored in the node. The node's average contribution is the average value of the coalitions with the node minus the average value of the coalitions without the node. We have also experimented with Shapley values, which are slightly more complex to compute, but the difference was negligible. \todo{Pevnak, could you add a reference to the paper where they did it the same way you mentioned?} 

An apparent disadvantage of using stochastically estimated values is lower \textbf{stability} of explanations, which is often desirable \cite{melis2018towards}. This can be mitigated by using a fixed seed to generate the subsets, but if stability is important, one of the non-stochastic heuristics may be a better choice.


\subsection{Sub-tree Selection Methods}\label{sec:ssp}
\todo{VL: I am not sure how to get rid of the "tree" here.}

The sub-tree selection problem is a direct generalization of the sub-set selection problem to the tree setting. The goal is to find a minimal sub-tree in the input tree, such that its evaluation by an expensive evaluation function is over the threshold $\tau$. For explanations, we will want a subset of an input sample that reaches threshold confidence of belonging to the correct class.
We first present the existing subset selection methods, mainly coming from feature selection~\citep{chandrashekar2014survey,li2017feature}, and then explain how to use them in the tree setting. Since the core problem addressed in this paper is explainability, we do not aspire to find the best possible sub-tree selection method, but rather to explore and compare a wide range of simple methods available in the literature. 

\subsubsection*{Subset selection problem}

Our variant of the subset selection problem for a set $N$ and an evaluation function $v$ is finding a subset $S \subseteq N$, such that $v(S)\geq \tau$ for a threshold $\tau$ and $|S|$ is minimized. Unfortunately, we generally cannot assume any structure of the evaluation function, such as monotonicity or submodularity. Exhaustive search is too expensive; hence, we use heuristics inspired by feature selection. The following procedures can be combined to create the right trade-off between the subset's quality and the computation speed.

\textbf{Greedy addition} is a simple and effective approximate solution technique for (approximately) submodular subset selection~\citep{das2011submodular}. It adds elements to $S$ one by one until the desired precision is achieved. Each new element maximizes the gain in the evaluation function over all elements that are not included in the subset yet.

\textbf{Heuristic addition} sorts the elements based on a heuristic ranking, such as the ones introduced in Section~\ref{sec:ranking}. It gradually adds elements one by one into an initially empty set from the most important ones until the threshold is reached.

\textbf{Random removal} (RR) can be run after any of the additions above. All the elements added in the first step are randomly shuffled and evaluated for removal. If any of them can be removed without the evaluation dropping below $\tau$, it is removed, and the next element in the sequence is evaluated. Once the algorithm reaches the end of the sequence, its starts over with a new permutation of remaining nodes until none of the nodes can be removed.

\textbf{Fine tuning} (FT) inspired by oscillating search \citep{somol2000oscillating} can further improve any initial set of elements. It adds $l$ elements greedily, such that the valuation function is maximized. Afterward, it repetitively removes the element that reduces the evaluation the least until no element can be removed without crossing the threshold. If this procedure removes the same elements it has just added, number $l$ is increased. For a current set $S \subset N$, we start with $l=1$ and end with $l=\min(5,2|S|)$.
Variations on this fine-tuning mechanism can be found in the literature (\eg, \citep{feige2011maximizing}), but since they do not provide formal guarantees without submodularity and the methods above achieve near-optimal results in our experiments, we do not evaluate them.

\subsubsection*{Applying subset selection in HMIL data}

When applying subset selection to find a sub-tree explaining a classifier's decision, the evaluation function is the output of the classifier (in the correct classification class). 
Since a data sample has a tree structure, not all subsets of nodes in the tree are meaningful explanations. 
We investigate three options for approaching this problem: one ignoring the structure and the other two that exploit it.

\textbf{Flat search} maintains the structure of the sample only for subset evaluation. Besides adding the root by default, it treats each node in the sample as a separate element and uses the subset selection algorithms from Section~\ref{sec:ssp}.  After finding the final subset, the nodes that are unreachable from the root are removed since they do not have any impact on classification.
Note that flat heuristic addition with ``GNN explainer mask ranking'' is the \textbf{GNN explainer}, which we implemented to exactly match  \citet{ying2019gnnexplainer}. The only difference is that instead of having the size of the explanation fixed as an input, we use \emph{the smallest size} that leads to a consistent explanation with the desired confidence. That means that the size of the explanation by GNN explainer cannot be further reduced by adapting its parameter $k$, without making the explanation belong to a different class than the original sample.

\textbf{Leaf search} runs the subset selection on all leaf nodes of the sample. If a leaf is included, all its predecessors are also included for model evaluation.

\textbf{Level-by-level search} 
searches for the explaining subset of the input hierarchical sample by optimizing one level of the tree at a time.
The root is always included.
On each level, it searches for a subset of children of the nodes included in the previous level using an algorithm from Section~\ref{sec:ssp}.
A child is considered to be present or absent with all of its descendants for model evaluation.
The descendants of the included children will be further pruned on the following levels.
The goal of the search on each level is to find the smallest subset of the children, which leads to an output of the model over threshold $\tau$. Since we selected such subset on the previous level with all descendants assumed to be present, a suitable subset must also exist on the next level. 
\section{Experiments}

We use HMIL neural networks for all our experiments (see Appendix~\ref{app:arch} for full details on the HMIL architecture). 
The neural network was trained using ADAM~\citep{adam} with minibatches of size 100 samples for 1000 training steps. 
We use binary classification into two classes and ensure that an empty JSON sample belongs to the negative class. 
The "confidence" is calculated as the difference between the output of positive and negative classes after softmax in the last layer of the neural networks. For computing GNN explainer and Banzhaf values, we use 200 steps/samples. The reported computation times correspond to a single thread on Intel(R) Xeon(R) Gold 5120 CPUs. See Appendix~\ref{app:hw} for details.

We first present thorough experiments on a wide range of smaller datasets with many repetitions. Afterward, we will show a use case where the size and speed of explanation are more crucial. The instructions on how to download the data and the source code are in Appendix~\ref{app:zip} and the instructions to re-create the experimental results are in Appendix~\ref{app:checklist}.

\subsection{Quantitative Analysis}
Following the experimental methodology for quantitative analysis introduced in \citet{ying2019gnnexplainer}, we generate synthetic data based on the existing datasets and deliberately include identifying \emph{concepts} (defined below) into the samples. We measure how well the explaining methods identify these concepts. 
To do that, we use real-world data from several domains in the JSON format: device properties obtained from a scan of a local network (\emph{deviceid}) \citep{deviceid}, symptoms and test results of hepatitis patients (\emph{hepatitis}) \citep{hepathitis}, and descriptions of molecules tested for mutagenicity on Salmonella typhimurium (\emph{mutagenesis}) \citep{mutagenesis}. 
For each dataset, we create a \textbf{schema} (we provide all schemata in Appendix~\ref{app:schemas}), and we use these schemata to generate realistic \textbf{synthetic data samples}.
Schemata, in addition to the structure of the data, carry statistics for each node, \eg, histograms of values or list lengths. It allows generating data samples with inserted concepts, while preserving most of the statistical properties of the original dataset. The generated dataset is referenced in Appendix~\ref{app:zip}.

A \textbf{concept} that decides whether a data sample is in a positive or a negative class is a set of sub-trees following the schema.
One sub-tree is composed from paths between the root and one of the value nodes in the tree. An example of a path in Fig.~\ref{fig:json} is {\scriptsize \mintinline{json}{{"upnp": [{"model_name" : "Sonos Play 3"}]}}}.
A set of paths is merged to form a sub-tree, \eg, a subtree with two leaves is {\scriptsize \mintinline[breaklines]{json}{{"upnp": [{"model_name" : "Sonos Play 3", "manufacturer": "Sonos, Inc."}]}}}.
A concept is then \textbf{a set of such sub-trees}.
A data sample is in the \textbf{positive class}, if and only if it includes at least one of the sub-trees in the concept completely. It typically includes much more data irrelevant to the classification. If a data sample is not in the positive class, it is in the \textbf{negative class}. For example, if a concept includes only one tree composed of two leaves, a sample containing only one of the leaves, but not the other, would be in the negative class. For each dataset, we generated 7 types of concepts: 
\begin{inparaenum}[(i)]
\item single path, 
\item two paths,
\item five paths, 
\item one tree composed of two paths, 
\item one tree composed of five paths, 
\item two trees composed of two paths each, and 
\item two trees composed of five paths.
\end{inparaenum}

Our synthetic dataset has limitations. For example, real-world problems are not perfectly separable and can allow multiple alternative explanations. However, removing these complications is exactly what allows us to rigorously compare to optimal explanations, which are not well-defined in fully realistic problems.

\subsubsection{Quantitative results}

 \begin{table*}[ht]
  \caption{Average number of excess leaves in the explanations and run time with standard error over all data sets. The prior work as reported and after hyper-parameter tuning is highlighted by the dashed line.
  }
  \centering
  \begin{tabular}{c|c|c|rr|rr|rr}
    \toprule
    \multicolumn{2}{c}{} & & \multicolumn{2}{c|}{Addition} & \multicolumn{2}{c|}{Add+RR} & \multicolumn{2}{c}{Add+RR+FT}\\
    \multicolumn{2}{c}{Search} & Ranking & Excess & Time [s] & Excess & Time [s] & Excess & Time [s]\\  

    \midrule
    \parbox[t]{2mm}{\multirow{7}{*}{\rotatebox[origin=c]{90}{Flat}}} 
 
   & Greedy    & -         & $0.79\pm .13$ & $0.37\pm .06$ & $0.07\pm .01$ & $0.28\pm .04$ & $0.05\pm .01$ & $0.69\pm .12$ \\
    \cmidrule(l){2-9}
    & \multirow{5}{*}{\rotatebox[origin=c]{90}{Heuristic}}  & GNN & $35.89\pm .76$ & $1.05\pm .01$ & $0.20\pm .02$ & $1.10\pm .01$ & $\mathbf{0.04}\pm .01$ & $1.47\pm .08$ \\
    & 			    & GNN2 & $2.39\pm .07$ & $1.03\pm .01$ & $0.10\pm .01$ & $1.03\pm .01$ & $\mathbf{0.04}\pm .01$ & $1.39\pm .08$ \\
    &           & Grad & $1.63\pm .23$ & $\mathbf{0.01}\pm .00$ & $0.10\pm .01$ & $\mathbf{0.01}\pm .00$ & $\mathbf{0.04}\pm .01$ & $0.43\pm .09$ \\
    &           & Banz & $\mathbf{0.34}\pm .04$ & $0.05\pm .00$ & $\mathbf{0.06}\pm .01$ & $0.05\pm .00$ & $\mathbf{0.04}\pm .01$ & $\mathbf{0.39}\pm .07$ \\
    &           & Rand & $22.04\pm .66$ & $0.03\pm .00$ & $0.15\pm .01$ & $0.06\pm .00$ & $\mathbf{0.04}\pm .01$ & $0.42\pm .08$ \\
    
    \midrule
    \parbox[t]{2mm}{\multirow{5}{*}{\rotatebox[origin=c]{90}{Leafs}}} &
     Greedy & - & $0.27\pm .08$ & $0.27\pm .22$ & $\mathbf{0.06}\pm .01$ & $0.23\pm .09$ & $\mathbf{0.06}\pm .01$ & $0.54\pm .11$ \\
    \cmidrule(l){2-9}
     & \multirow{3}{*}{\rotatebox[origin=c]{90}{Heur.}} 
& Grad &$0.50\pm .06$ & $\mathbf{0.02}\pm .00$ & $0.10\pm .01$ & $\mathbf{0.03}\pm .00$ & $0.08\pm .01$ & $0.54\pm .15$ \\
& & Banz &$\mathbf{0.42}\pm .06$ & $0.17\pm .01$ & $0.09\pm .01$ & $0.16\pm .01$ & $0.07\pm .01$ & $\mathbf{0.53}\pm .14$ \\
& & Rand &$31.44\pm .70$ & $0.06\pm .00$ & $0.26\pm .02$ & $0.19\pm .01$ & $0.10\pm .01$ & $0.77\pm .16$ \\
    
    \midrule
    \parbox[t]{2mm}{\multirow{6}{*}{\rotatebox[origin=c]{90}{Level-by-level}}}  
    & Greedy    & -        & $0.23\pm .04$ & $0.03\pm .01$ & $\mathbf{0.09}\pm .01$ & $0.05\pm .01$ & $0.09\pm .01$ & $\mathbf{0.03}\pm .01$ \\
    \cmidrule(l){2-9}
    & \multirow{5}{*}{\rotatebox[origin=c]{90}{Heuristic}}  & GNN & $6.07\pm .28$ & $1.04\pm .01$ & $0.30\pm .02$ & $1.04\pm .01$ & $0.24\pm .02$ & $1.07\pm .01$ \\
    &           & GNN2 & $1.79\pm .05$ & $1.04\pm .01$ & $0.12\pm .01$ & $1.03\pm .01$ & $0.12\pm .01$ & $1.05\pm .01$ \\
    &           & Grad & $0.42\pm .04$ & $\mathbf{0.01}\pm .00$ & $0.17\pm .01$ & $\mathbf{0.01}\pm .00$ & $0.16\pm .01$ & $0.04\pm .00$ \\
    &           & Banz & $\mathbf{0.22}\pm .02$ & $0.05\pm .00$ & $\mathbf{0.09}\pm .01$ & $0.05\pm .00$ & $\mathbf{0.08}\pm .01$ & $0.07\pm .00$ \\
    &           & Rand & $4.39\pm .19$ & $\mathbf{0.01}\pm .00$ & $0.28\pm .02$ & $\mathbf{0.01}\pm .00$ & $0.22\pm .02$ & $0.04\pm .00$ \\
    \bottomrule
  \end{tabular}
  \begin{tikzpicture}[overlay]
		\draw[dashed,red,thick](-9.4,2.55) -- (-6.3,2.55)  -- (-6.3,1.8) -- (-9.4,1.8) -- cycle;
  \end{tikzpicture}
  \label{tab:synthetic_results}
\end{table*}

For each of 3 domains and 7 concept variants inserted in data, we have generated 20 datasets of 20,000 samples to train classifiers. 
We have randomly selected 1,000 samples and explained them using all combinations of ranking methods (Section~\ref{sec:ranking}) and selection methods~(Section~\ref{sec:ssp}). The confidence threshold $\tau$ in the explanation was set to $0.9$ times the confidence on the full sample since explanations created with a low threshold are noisy and frequently miss the concepts.

Table~\ref{tab:synthetic_results} shows aggregated results from all generated datasets.
We measure the performance by (1) the number of \emph{excess leaves} in the tree that are part of the explanation while they are not part of the inserted concept, and (2) computation time required to find the explanation of a single sample. The number of excess leaves measures how well we achieve the goal of finding the minimal possible explanation, and hence how faithful and comprehensible the explanations are (see Section~\ref{sec:intro}). ``Rand'' denotes a baseline of random ranking.

We first focus on the first column in the table (``Addition''), which compares just adding features based on various heuristics until a subset consistent with the whole sample classification is found. In the ``Flat'' case, which does not explicitly exploit the tree structure, Banzhaf values are the most effective heuristic leading to only $0.34$ excess leaves. This is expected since this heuristic most explicitly measures the importance of including vs. not including a node. GNN explainer does not perform well in HMIL data. Even the simple model gradient heuristic creates a surplus of 1.63 leaves as opposed to 2.39 in the case of GNN explainer\footnote{GNN explainer's explanation size cannot be further reduced without losing explanation consistency (see Section~\ref{sec:ssp}).}, even after tuning its hyper-parameters (denoted GNN2). The ordering of the heuristics stays the same, even if we add the nodes ``Level-by-level''. However, the excess for all methods is reduced. A notable improvement can be seen in the performance of ``Greedy''. While in ``Flat'', each node's contribution is evaluated separately, in ``Level-by-level'', a node on a particular level is evaluated as if the whole sub-tree below the node was included in the sample. All ``Level-by-level'' methods produced shorter explanations in similar or substantially shorter times, with Banzhaf heuristic being the most effective.

In the second column of Table~\ref{tab:synthetic_results} (``Add+RR''), the addition is followed by the random removal (RR) procedure. It substantially reduces the excess of all methods, while the increase in the computation time is insignificant with better heuristics. The number of leaves before running RR is already small, so the extra cost is only a few more evaluations of the model. Even after RR, Banzhaf values lead to the smallest excess; however, they are matched by greedy methods. 

The third column of the table adds the fine-tuning (FT) to the previous two steps. It is a computationally expensive procedure. In the ``Flat'' approach, it may increase the time required for explanation over 40 times (Grad). However, since it can undo any errors caused by the heuristics, it eventually always reached the overall minimum of $0.04$ excess leaves in the explanations. The computational overhead is much less severe in the ``Level-by-level'' approach, because the sizes of the sets explored on each level are smaller and further restricted by the dependence on the nodes fixed in the previous level. These restrictions also prevent reaching minimal explanations. The best excess ($0.08$) is again reached by Banzhaf values, while the average computation time increased from $0.05$ to $0.07$. The fastest method is greedy, since it is using the increments in the model output already in addition, and hence FT is less likely to find further improvements based on the same measure.

In summary, Banzhaf values are the most effective heuristic for guiding the simpler search methods. If we can afford a more expensive search, the difference between the heuristics is reduced. If we care about computation time, we should definitely use the ``Level-by-level'' approaches we introduce, either with Banzhaf values or Greedy addition. 
Detailed results with the break-downs based on the sought concept's complexity are in Appendix~\ref{app:exp}. The relative performance differences among the algorithms are very consistent with the overall results.

\subsection{Qualitative Analysis}

\begin{table*}[th!]
  \caption{Characteristics of explanations on large data samples from Cuckoo Sandbox. $^*$Statistics from less samples (43) due to computational complexity.} \label{tab:cuckoo}
  \centering
 \scriptsize{
\begin{tabular}{lrrrrr}
\toprule
Method & Time [s] & ~~\#Gradients & ~~\#Inferences  & ~~Explanation size & ~~Input size \\
\midrule
GNN Explainer (tuned) & 413 & 200 & 984 & 70.24 & $9.9 \cdot 10^4$ \\
Leafs-Greedy+RR$^*$ & 8375 & 0 & 72064 & 1.00 & $3.8\cdot 10^4$ \\
Leafs-banz-Add+RR & 2689 & 0 & 11554 & 5.07 & $9.9 \cdot 10^4$ \\
Leafs-grad-Add+RR & 3913 & 1 & 19364 & 9.49 & $9.9 \cdot 10^4$ \\\midrule 
LbyL-GNN-Add+RR & 731 & 200 & 1715 & 4.47 & $9.9 \cdot 10^4$ \\
LbyL-Greedy+RR & 500 & 0 & 170459 & 1.16 & $9.9 \cdot 10^4$ \\
LbyL-banz-Add+RR & 101 & 0 & 595 & 1.33 & $9.9 \cdot 10^4$ \\
LbyL-grad-Add+RR & 38 & 1 & 2686 & 1.56 & $9.9 \cdot 10^4$ \\ \bottomrule
\end{tabular}}
\end{table*}

\begin{figure}[th]
\begin{minipage}{0.5\columnwidth}
\begin{subfigure}{\columnwidth}
\begin{minted}[fontsize=\tiny]{json}
{"upnp": [
  { "device_type": 
      "urn:schemas-upnp-org:device:DigitalSecurityCamera:1",
    "model_description": 
      "H.264 MegaPixel NetworkCamera[Wireless]",
    "services": [ "urn:upnp-org:serviceId:dummy1" ]}]}
\end{minted}
\caption{Flat-Grad-Add+RR+FT}\label{fig:our1}
\end{subfigure}
\begin{subfigure}{\columnwidth}
\begin{minted}[fontsize=\tiny]{json}
{
  "services": [ { "port": [ 80, 5001]} ],
  "mdns_services": [ "_hap._tcp.local."]}
\end{minted}
\caption{Flat-Grad-Add+RR+FT}\label{fig:our2}
\end{subfigure}
\end{minipage}
\begin{minipage}{0.49\columnwidth}
\begin{subfigure}{\columnwidth}
\begin{minted}[fontsize=\tiny]{json}
{"ssdp": [
  { "nt": ["upnp:rootdevice",
      "urn:schemas-upnp-org:device:DigitalSecurityCamera:1"],
    "server": 
      "Linux/2.6.35.6-45.fc14.i686 UPnP/1.0 miniupnpd/1.0"}],
 "upnp": [
  { "model_name": "Sentry H.130",
    "manufacturer": "NetworkCamera",
    "device_type": 
      "urn:schemas-upnp-org:device:DigitalSecurityCamera:1",
    "model_description": 
      "H.264 MegaPixel NetworkCamera[Wireless]",
    "services": ["urn:upnp-org:serviceId:dummy1"]}],
  "services": [{ "port": [80, 1900],
                 "protocol": ["tcp", "udp"]}]}
\end{minted}
\caption{GNN explainer}\label{fig:gnn1}
\end{subfigure}
\end{minipage}
\caption{Explanation of the ``security camera'' classification in the device identification domain:  (a-b) explanations of two different samples by our method; (c) explanation of the same sample as in (a) by GNN explainer.}\label{fig:explanations}
\end{figure}

Fig.~\ref{fig:explanations} shows example explanations of two samples in the device identification problem trained on the real-world (not synthetic) data. Fig.~\ref{fig:our1} and \ref{fig:our2} are explanations of two different samples from the ``security camera'' class computed by our method. The explanations are compact and focus on various keys in the JSON file. ``hap'' in the second sample stands for "HomeKit Accessory Protocol". Fig.~\ref{fig:gnn1} is an explanation of the same sample as in Fig.~\ref{fig:our1} by GNN explainer. It adds many more sub-trees to the sample to ensure the correct classification.
\todo[inline]{VL: Can we find a more impressive example? HAP+80+5001 can easily be a smart speaker, NAS, media center. I assume this explanation is valid mainly because there are no other protocols, which would indicate other classes.}

Having a quick algorithm that provides compact explanations is important, mainly for debugging classifiers working with large raw data samples. 
For example, dynamic analysis reports of executable files from Cuckoo Sandbox available from \citep{rcc-dataset} are between few kilobytes and one gigabyte of data, with the largest reports including over fifty million nodes.
We used them to train an HMIL classifier distinguishing malware and clean executables.
Since the file names in the reports include the correct class, we replaced them with an uninformative token before the training.
Table~\ref{tab:cuckoo} summarizes the explanations of 60 random samples by various methods. We report the number of inferences of the model while computing the explanation, the number of gradient computations on the model, the number of leaves in the explanation, and the average number of nodes in the full sample.
Not only is the GNN explainer over 10 times slower than our fastest method due to 200 computations of the model gradient, but it also produces explanations with over 70 leaves on average, which gives very little insight into the classifier. The general patterns of the speed and explanation size follow the synthetic data experiments. The ``Level-by-level'' methods are generally at least an order of magnitude faster less precise. A notable exception is the Leafs-Greedy+RR method, which always found an explanation with only a single leaf, but took over 200 times longer than the fastest method.  
The short explanations by our methods generally included only one leaf, such as, {\scriptsize \mintinline{json}{{"info" : { "machine" : { "started_on" : "2019-07-18 18:41:34" }}}}}. This helped us understand that the malware and clean samples had been collected in distinct time periods, leading to an incorrectly learned model. This led us to take corrective action and normalize all time-related fields in the report. The consequently trained model provided meaningful explanations.

\section{Conclusions}

We investigate the explainability of classifiers applied to raw hierarchically structured data.
We argue for the minimal input subset, classified to the same class as the whole input, as a suitable form of explanations for models on this data.
We show how existing subset selection methods can be adapted for finding explanations and explore a set of heuristics to guide them.
In contrast to general graph neural networks, the search for minimal subsets is well-guided even by simple gradients, but the best results are achieved using the greedy search or Banzhaf values.
We propose an extension of subset search methods that proceeds level-by-level in the hierarchical data and show it brings substantial computational savings with a minimal impact on the explanation size.

Besides the rigorous experiments on semi-synthetic data, we provide a qualitative analysis of the best methods on large-scale real-world data and demonstrate how the produced explanations help identify problems in raw datasets, such as the misleading timestamps. We can explain large input samples with millions of nodes order of magnitude faster than GNN explainer, and our explanations are far more compact. Our methods are not sensitive to hyper-parameter settings, and since Banzhaf values computation does not require computation of gradients, they are also applicable for non-differentiable models.

\bibliographystyle{splncs04}
\bibliography{refs}


\appendix

\newpage
\clearpage
\appendix
\section{Library and data to recreate experiments}\label{app:zip}
The complete source code and raw experiment results are provided in data appendix. The experimental setup with full datasets, trained models, and collected results is available at
\href{https://drive.google.com/file/d/1xyx29yUwaw0Zqr1T5X0JSvk1MJmgMe6q/view?usp=sharing}{\tiny https://drive.google.com/file/d/1xyx29yUwaw0Zqr1T5X0JSvk1MJmgMe6q}. Note that it is 3.5Gb packed and 36Gb unpacked.

\section{HMIL NN architecture and training}\label{app:arch}

The HMIL neural network model is automatically constructed from the schema of the problem. Let $n$ be a node in the data sample, $sch(n)$ be the corresponding node of the schema, $succ(n)$ be the successor nodes of $n$ in the sample and $c \in succ(n)$ one of the successors. We denote by $\phi^{sch(n),sch(c)}$ the embedding and $\agg^{sch(n)}$ the aggregation functions used in node $n$. Note that we use $\agg$ as an operator, similar to $\sum$ or $\prod$, and indicate the elements it aggregates over below it. The HMIL neural network realizes the following recursively defined function:
\begin{equation}\label{eq:hmil}
h(n) = \agg^{sch(n)}_{c \in succ(n)} \phi^{sch(n),sch(c)}\left(h(c)\right)
\end{equation}
Function $h$ in atomic nodes encodes values of any type into vectors of real numbers.  If the value is a number it is used as it is. If a value is a string occurring less than 100 times, then the string is treated as a categorical variable and encoded by one-hot encoding, else the strings are represented as histograms of a trigrams compacted to 2053 buckets. For a fixed ordering of trigrams, the number of occurrences of the $i$th trigram is added to the bucket number $i \mod 2053$.
The embedding functions in the inner nodes are usually feed-forward neural networks. The aggregation functions may be element-wise max/mean, concatenation of the successors' embeddings, but also an LSTM or any other differentiable machine learning model. For this paper, we assume that the aggregation function can aggregate any subset of successor embeddings.

Since we work with the inputs in the JSON format, there are two types of inner nodes in the tree: dictionary nodes and list nodes (see the main text). A \textbf{dictionary node} $n$ usually has a small fixed number of keys. In the schema, all the keys in the nodes corresponding to $sch(n)$ are collected and sorted. The aggregation function $\agg^{sch(n)}$ concatenates the embedding of the children corresponding to these keys in the fixed order. If some of the children are missing, a default vector of values is used instead of the embedding. This default vectors, separate for each child in each dictionary node, are optimized as additional parameters in training the HMIL neural network. The final output of the aggregation is than constructed by single layer of $k$ ReLU units. A \textbf{list node} may have an arbitrary number of successors. The aggregation function computes coordinate-wise maximum and mean over all successors' embeddings. The mean and maximum vectors are concatenated and followed by a single layer of $k$ ReLU units. All embedding functions inside the HMIL network were also single layer of $k$ ReLU units.

Inside the network, all embedding functions, all aggregation functions $\agg$, and all concatenation in dictionary nodes were followed by a single dense layer with $k=5$ units and with \texttt{relu} non-linearity. In few cases, where $k=5$ was not sufficient, we have used $k=10.$ Only the output feed-forward neural networks contained two layers, where the first layer was as above and the second was linear with two units, because all problems we are solving are binary.

All layers were intialized randomly using glorot scheme. The neural network was trained using ADAM with default settings with randonly sampled minibatches of size 100 samples for 2000 training steps. Importantly, since we wanted to ensure that concepts were classified correctly, we have trained 10 neural networks and selected the one which (i) classified empty sample (JSON) to negative class and (ii) classified all concepts with highest average "confidence", which was typically above 0.9 after the training. These criteria have ensured that the network has correctly learned concepts and we will not be explaining noise in the data. The "confidence" is calculated as difference between output of positive and negative class after softmax.

\section{Detailed experimental results}\label{app:exp}
Tables below shows average excess of leaves in the explanation and average explanation times on individual problems. For each task, averages are over 20 versions of three problems (20 * 3 experiments).

\subsection{one of 1 1trees}
  \resizebox{\columnwidth}{!}{
\begin{tabular}{c|@{~}c@{~}|@{~}c@{~}|rr|rr|rr}
  \toprule
  \multicolumn{2}{c}{} & & \multicolumn{2}{c|}{Addition} & \multicolumn{2}{c|}{Add+RR} & \multicolumn{2}{c}{Add+RR+FT}\\
  \multicolumn{2}{c}{Search} & Rank. & Excess & Time [s] & Excess & Time [s] & Excess & Time [s]\\  
  \midrule
  \parbox[t]{2mm}{\multirow{6}{*}{\rotatebox[origin=c]{90}{Flat}}} 
 & Greedy    & -         & $1.06\pm .17$ & $1.70\pm .85$ & $0.05\pm .02$ & $1.79\pm .94$ & $0.05\pm .02$ & $2.96\pm .97$ \\
  \cmidrule(l){2-9}
  & \multirow{5}{*}{\rotatebox[origin=c]{90}{Heuristic}}  & GNN & $21.91\pm .32$ & $1.43\pm .03$ & $0.07\pm .02$ & $1.46\pm .04$ & $0.00\pm .00$ & $2.29\pm .83$ \\
  &           & GNN2 & $5.37\pm .42$ & $2.53\pm .09$ & $0.05\pm .02$ & $2.50\pm .09$ & $0.01\pm .01$ & $3.46\pm .86$ \\
  &           & Grad & $0.07\pm .03$ & $0.01\pm .00$ & $0.02\pm .01$ & $0.01\pm .00$ & $0.01\pm .01$ & $0.72\pm .77$ \\
  &           & Banz & $0.41\pm .19$ & $0.11\pm .01$ & $0.00\pm .00$ & $0.11\pm .01$ & $0.00\pm .00$ & $0.91\pm .83$ \\
  &           & Rnd & $22.49\pm .46$ & $0.06\pm .01$ & $0.06\pm .02$ & $0.12\pm .02$ & $0.00\pm .00$ & $0.93\pm .81$ \\
  \midrule
  \parbox[t]{2mm}{\multirow{6}{*}{\rotatebox[origin=c]{90}{Level-by-level}}}  
  & Greedy    & -        & $0.01\pm .01$ & $0.05\pm .01$ & $0.01\pm .01$ & $0.05\pm .01$ & $0.01\pm .01$ & $0.10\pm .02$ \\
  \cmidrule(l){2-9}
  & \multirow{5}{*}{\rotatebox[origin=c]{90}{Heuristic}}  GNN & $5.95\pm .72$ & $1.36\pm .03$ & $0.24\pm .05$ & $1.40\pm .03$ & $0.00\pm .00$ & $2.32\pm .11$ \\
  &           & GNN2 & $3.25\pm .21$ & $2.57\pm .09$ & $0.07\pm .02$ & $2.47\pm .08$ & $0.06\pm .02$ & $2.60\pm .09$ \\
  &           & Grad & $0.05\pm .02$ & $0.02\pm .00$ & $0.04\pm .01$ & $0.02\pm .00$ & $0.02\pm .01$ & $1.09\pm .12$ \\
  &           & Banz & $0.07\pm .02$ & $0.11\pm .01$ & $0.00\pm .00$ & $0.11\pm .01$ & $0.00\pm .00$ & $0.90\pm .73$ \\
  &           & Rnd & $5.29\pm .63$ & $0.02\pm .00$ & $0.25\pm .05$ & $0.02\pm .00$ & $0.00\pm .00$ & $1.02\pm .09$ \\\hline\hline
  \bottomrule
\end{tabular}}

\subsection{one of 1 2trees}
  \resizebox{\columnwidth}{!}{
\begin{tabular}{c|@{~}c@{~}|@{~}c@{~}|rr|rr|rr}
  \toprule
  \multicolumn{2}{c}{} & & \multicolumn{2}{c|}{Addition} & \multicolumn{2}{c|}{Add+RR} & \multicolumn{2}{c}{Add+RR+FT}\\
  \multicolumn{2}{c}{Search} & Rank. & Excess & Time [s] & Excess & Time [s] & Excess & Time [s]\\  
  \midrule
  \parbox[t]{2mm}{\multirow{6}{*}{\rotatebox[origin=c]{90}{Flat}}} 
 & Greedy    & -         & $0.71\pm .10$ & $1.32\pm .62$ & $0.09\pm .03$ & $1.37\pm .67$ & $0.03\pm .01$ & $1.82\pm .76$ \\
  \cmidrule(l){2-9}
  & \multirow{5}{*}{\rotatebox[origin=c]{90}{Heuristic}}  & GNN & $26.11\pm .33$ & $1.43\pm .03$ & $0.03\pm .01$ & $1.52\pm .04$ & $0.00\pm .00$ & $1.83\pm .12$ \\
  &           & GNN2 & $4.47\pm .35$ & $2.37\pm .07$ & $0.02\pm .01$ & $2.53\pm .08$ & $0.00\pm .00$ & $2.82\pm .16$ \\
  &           & Grad & $0.09\pm .02$ & $0.01\pm .00$ & $0.00\pm .00$ & $0.01\pm .00$ & $0.00\pm .00$ & $0.29\pm .06$ \\
  &           & Banz & $0.18\pm .08$ & $0.10\pm .01$ & $0.00\pm .00$ & $0.11\pm .01$ & $0.00\pm .00$ & $0.38\pm .06$ \\
  &           & Rnd & $26.85\pm .51$ & $0.08\pm .01$ & $0.03\pm .01$ & $0.13\pm .02$ & $0.00\pm .00$ & $0.46\pm .10$ \\
  \midrule
  \parbox[t]{2mm}{\multirow{6}{*}{\rotatebox[origin=c]{90}{Level-by-level}}}  
  & Greedy    & -        & $0.03\pm .01$ & $0.07\pm .02$ & $0.02\pm .01$ & $0.07\pm .02$ & $0.02\pm .01$ & $0.14\pm .04$ \\
  \cmidrule(l){2-9}
  & \multirow{5}{*}{\rotatebox[origin=c]{90}{Heuristic}}  GNN & $7.24\pm .75$ & $1.45\pm .04$ & $0.05\pm .02$ & $1.44\pm .04$ & $0.00\pm .00$ & $1.69\pm .12$ \\
  &           & GNN2 & $2.61\pm .19$ & $2.40\pm .07$ & $0.03\pm .01$ & $2.45\pm .08$ & $0.01\pm .01$ & $2.62\pm .10$ \\
  &           & Grad & $0.08\pm .02$ & $0.02\pm .00$ & $0.01\pm .01$ & $0.02\pm .00$ & $0.00\pm .00$ & $0.48\pm .11$ \\
  &           & Banz & $0.11\pm .07$ & $0.11\pm .01$ & $0.02\pm .01$ & $0.11\pm .01$ & $0.00\pm .00$ & $0.43\pm .09$ \\
  &           & Rnd & $6.75\pm .69$ & $0.02\pm .00$ & $0.05\pm .02$ & $0.02\pm .00$ & $0.00\pm .00$ & $0.40\pm .09$ \\
  \bottomrule
\end{tabular}}

\subsection{one of 1 5trees}
  \resizebox{\columnwidth}{!}{
\begin{tabular}{c|@{~}c@{~}|@{~}c@{~}|rr|rr|rr}
  \toprule
  \multicolumn{2}{c}{} & & \multicolumn{2}{c|}{Addition} & \multicolumn{2}{c|}{Add+RR} & \multicolumn{2}{c}{Add+RR+FT}\\
  \multicolumn{2}{c}{Search} & Rank. & Excess & Time [s] & Excess & Time [s] & Excess & Time [s]\\  
  \midrule
  \parbox[t]{2mm}{\multirow{6}{*}{\rotatebox[origin=c]{90}{Flat}}} 
 & Greedy    & -         & $0.76\pm .10$ & $1.13\pm .38$ & $0.09\pm .02$ & $1.11\pm .35$ & $0.07\pm .02$ & $2.24\pm .72$ \\
  \cmidrule(l){2-9}
  & \multirow{5}{*}{\rotatebox[origin=c]{90}{Heuristic}}  & GNN & $30.81\pm .46$ & $1.44\pm .03$ & $0.17\pm .03$ & $1.50\pm .04$ & $0.02\pm .01$ & $2.70\pm .46$ \\
  &           & GNN2 & $4.03\pm .31$ & $2.59\pm .08$ & $0.14\pm .02$ & $2.58\pm .09$ & $0.03\pm .01$ & $4.25\pm .72$ \\
  &           & Grad & $0.41\pm .07$ & $0.01\pm .00$ & $0.15\pm .02$ & $0.02\pm .00$ & $0.01\pm .00$ & $0.75\pm .19$ \\
  &           & Banz & $0.49\pm .10$ & $0.11\pm .01$ & $0.09\pm .02$ & $0.11\pm .01$ & $0.01\pm .01$ & $0.83\pm .19$ \\
  &           & Rnd & $31.58\pm .66$ & $0.08\pm .01$ & $0.16\pm .03$ & $0.15\pm .02$ & $0.02\pm .01$ & $1.27\pm .68$ \\
  \midrule
  \parbox[t]{2mm}{\multirow{6}{*}{\rotatebox[origin=c]{90}{Level-by-level}}}  
  & Greedy    & -        & $0.18\pm .05$ & $0.10\pm .04$ & $0.10\pm .03$ & $0.10\pm .04$ & $0.10\pm .03$ & $0.19\pm .06$ \\
  \cmidrule(l){2-9}
  & \multirow{5}{*}{\rotatebox[origin=c]{90}{Heuristic}}  GNN & $10.01\pm .86$ & $1.46\pm .04$ & $0.29\pm .04$ & $1.41\pm .03$ & $0.03\pm .01$ & $2.30\pm .35$ \\
  &           & GNN2 & $2.67\pm .20$ & $2.62\pm .08$ & $0.19\pm .03$ & $2.59\pm .08$ & $0.18\pm .03$ & $2.70\pm .09$ \\
  &           & Grad & $0.31\pm .05$ & $0.02\pm .00$ & $0.29\pm .04$ & $0.02\pm .00$ & $0.02\pm .01$ & $0.93\pm .21$ \\
  &           & Banz & $0.33\pm .08$ & $0.11\pm .01$ & $0.14\pm .03$ & $0.12\pm .01$ & $0.02\pm .01$ & $0.86\pm .20$ \\
  &           & Rnd & $9.44\pm .82$ & $0.03\pm .00$ & $0.28\pm .04$ & $0.03\pm .00$ & $0.03\pm .01$ & $0.98\pm .29$ \\
  \bottomrule
\end{tabular}}

\subsection{one of 2 5trees}
  \resizebox{\columnwidth}{!}{
\begin{tabular}{c|@{~}c@{~}|@{~}c@{~}|rr|rr|rr}
  \toprule
  \multicolumn{2}{c}{} & & \multicolumn{2}{c|}{Addition} & \multicolumn{2}{c|}{Add+RR} & \multicolumn{2}{c}{Add+RR+FT}\\
  \multicolumn{2}{c}{Search} & Rank. & Excess & Time [s] & Excess & Time [s] & Excess & Time [s]\\  
  \midrule
  \parbox[t]{2mm}{\multirow{6}{*}{\rotatebox[origin=c]{90}{Flat}}} 
 & Greedy    & -         & $2.11\pm .57$ & $1.10\pm .31$ & $0.15\pm .03$ & $1.10\pm .30$ & $0.15\pm .03$ & $2.77\pm .82$ \\
  \cmidrule(l){2-9}
  & \multirow{5}{*}{\rotatebox[origin=c]{90}{Heuristic}}  & GNN & $27.64\pm .41$ & $1.41\pm .03$ & $0.23\pm .03$ & $1.47\pm .03$ & $0.15\pm .03$ & $3.52\pm .71$ \\
  &           & GNN2 & $3.85\pm .27$ & $2.59\pm .10$ & $0.26\pm .03$ & $2.59\pm .09$ & $0.15\pm .03$ & $5.06\pm .85$ \\
  &           & Grad & $0.71\pm .24$ & $0.01\pm .00$ & $0.20\pm .03$ & $0.02\pm .00$ & $0.14\pm .02$ & $1.82\pm .63$ \\
  &           & Banz & $0.40\pm .08$ & $0.11\pm .01$ & $0.16\pm .03$ & $0.12\pm .01$ & $0.14\pm .03$ & $2.06\pm .55$ \\
  &           & Rnd & $29.99\pm .63$ & $0.07\pm .01$ & $0.23\pm .03$ & $0.15\pm .02$ & $0.15\pm .03$ & $2.10\pm .63$ \\
  \midrule
  \parbox[t]{2mm}{\multirow{6}{*}{\rotatebox[origin=c]{90}{Level-by-level}}}  
  & Greedy    & -        & $0.59\pm .16$ & $0.10\pm .04$ & $0.20\pm .03$ & $0.07\pm .02$ & $0.20\pm .03$ & $0.16\pm .05$ \\
  \cmidrule(l){2-9}
  & \multirow{5}{*}{\rotatebox[origin=c]{90}{Heuristic}}  GNN & $6.81\pm .65$ & $1.39\pm .03$ & $0.33\pm .04$ & $1.37\pm .03$ & $0.15\pm .03$ & $3.02\pm .50$ \\
  &           & GNN2 & $2.25\pm .17$ & $2.55\pm .09$ & $0.28\pm .03$ & $2.59\pm .09$ & $0.26\pm .03$ & $2.79\pm .12$ \\
  &           & Grad & $0.52\pm .15$ & $0.02\pm .00$ & $0.27\pm .04$ & $0.02\pm .00$ & $0.15\pm .03$ & $1.63\pm .46$ \\
  &           & Banz & $0.29\pm .06$ & $0.11\pm .01$ & $0.19\pm .03$ & $0.11\pm .01$ & $0.15\pm .03$ & $1.79\pm .47$ \\
  &           & Rnd & $7.10\pm .67$ & $0.02\pm .00$ & $0.33\pm .04$ & $0.03\pm .00$ & $0.15\pm .03$ & $1.76\pm .48$ \\
  \bottomrule
\end{tabular}}

\subsection{one of 2 1trees}
  \resizebox{\columnwidth}{!}{
\begin{tabular}{c|@{~}c@{~}|@{~}c@{~}|rr|rr|rr}
  \toprule
  \multicolumn{2}{c}{} & & \multicolumn{2}{c|}{Addition} & \multicolumn{2}{c|}{Add+RR} & \multicolumn{2}{c}{Add+RR+FT}\\
  \multicolumn{2}{c}{Search} & Rank. & Excess & Time [s] & Excess & Time [s] & Excess & Time [s]\\  
  \midrule
  \parbox[t]{2mm}{\multirow{6}{*}{\rotatebox[origin=c]{90}{Flat}}} 
 & Greedy    & -         & $0.49\pm .10$ & $0.97\pm .57$ & $0.09\pm .02$ & $0.97\pm .58$ & $0.08\pm .02$ & $1.27\pm .67$ \\
  \cmidrule(l){2-9}
  & \multirow{5}{*}{\rotatebox[origin=c]{90}{Heuristic}}  & GNN & $13.44\pm .06$ & $1.39\pm .03$ & $0.17\pm .03$ & $1.39\pm .03$ & $0.01\pm .01$ & $1.85\pm .10$ \\
  &           & GNN2 & $4.17\pm .26$ & $2.74\pm .10$ & $0.04\pm .02$ & $2.79\pm .10$ & $0.01\pm .01$ & $3.15\pm .19$ \\
  &           & Grad & $2.37\pm .60$ & $0.02\pm .00$ & $0.03\pm .01$ & $0.03\pm .00$ & $0.00\pm .00$ & $0.36\pm .08$ \\
  &           & Banz & $0.21\pm .05$ & $0.11\pm .01$ & $0.00\pm .00$ & $0.11\pm .01$ & $0.00\pm .00$ & $0.46\pm .09$ \\
  &           & Rnd & $13.02\pm .15$ & $0.04\pm .01$ & $0.16\pm .03$ & $0.08\pm .01$ & $0.01\pm .01$ & $0.51\pm .09$ \\
  \midrule
  \parbox[t]{2mm}{\multirow{6}{*}{\rotatebox[origin=c]{90}{Level-by-level}}}  
  & Greedy    & -        & $0.06\pm .01$ & $0.05\pm .02$ & $0.06\pm .01$ & $0.05\pm .01$ & $0.06\pm .01$ & $0.12\pm .04$ \\
  \cmidrule(l){2-9}
  & \multirow{5}{*}{\rotatebox[origin=c]{90}{Heuristic}}  GNN & $4.43\pm .52$ & $1.37\pm .03$ & $0.29\pm .05$ & $1.37\pm .03$ & $0.02\pm .01$ & $1.71\pm .10$ \\
  &           & GNN2 & $2.47\pm .18$ & $2.62\pm .09$ & $0.07\pm .02$ & $2.67\pm .10$ & $0.07\pm .02$ & $2.72\pm .12$ \\
  &           & Grad & $0.87\pm .24$ & $0.02\pm .00$ & $0.06\pm .02$ & $0.02\pm .00$ & $0.00\pm .00$ & $0.49\pm .09$ \\
  &           & Banz & $0.14\pm .03$ & $0.11\pm .01$ & $0.01\pm .01$ & $0.12\pm .01$ & $0.00\pm .00$ & $0.46\pm .08$ \\
  &           & Rnd & $3.85\pm .47$ & $0.02\pm .00$ & $0.29\pm .05$ & $0.02\pm .00$ & $0.02\pm .01$ & $0.49\pm .08$ \\\hline\hline
  \bottomrule
\end{tabular}}

\subsection{one of 5 1trees}
  \resizebox{\columnwidth}{!}{
\begin{tabular}{c|@{~}c@{~}|@{~}c@{~}|rr|rr|rr}
  \toprule
  \multicolumn{2}{c}{} & & \multicolumn{2}{c|}{Addition} & \multicolumn{2}{c|}{Add+RR} & \multicolumn{2}{c}{Add+RR+FT}\\
  \multicolumn{2}{c}{Search} & Rank. & Excess & Time [s] & Excess & Time [s] & Excess & Time [s]\\  
  \midrule
  \parbox[t]{2mm}{\multirow{6}{*}{\rotatebox[origin=c]{90}{Flat}}} 

 & Greedy    & -         & $0.60\pm .19$ & $1.12\pm .32$ & $0.02\pm .01$ & $1.11\pm .31$ & $0.02\pm .01$ & $1.57\pm .75$ \\
  \cmidrule(l){2-9}
  & \multirow{5}{*}{\rotatebox[origin=c]{90}{Heuristic}}  & GNN & $9.64\pm .04$ & $1.26\pm .04$ & $0.07\pm .02$ & $1.28\pm .03$ & $0.01\pm .00$ & $1.91\pm .99$ \\
  & 			    & GNN2 & $3.54\pm .26$ & $2.22\pm .09$ & $0.03\pm .01$ & $2.20\pm .09$ & $0.01\pm .00$ & $3.29\pm .64$ \\
  &           & Grad & $3.07\pm .71$ & $0.02\pm .00$ & $0.06\pm .02$ & $0.03\pm .00$ & $0.00\pm .00$ & $0.51\pm .62$ \\
  &           & Banz & $0.34\pm .08$ & $0.10\pm .02$ & $0.01\pm .01$ & $0.11\pm .02$ & $0.00\pm .00$ & $0.54\pm .67$ \\
  &           & Rnd & $8.82\pm .08$ & $0.05\pm .08$ & $0.05\pm .02$ & $0.07\pm .03$ & $0.01\pm .00$ & $1.50\pm .93$ \\
  \midrule
  \parbox[t]{2mm}{\multirow{6}{*}{\rotatebox[origin=c]{90}{Level-by-level}}}  
  & Greedy    & -        & $0.00\pm .00$ & $0.06\pm .04$ & $0.00\pm .00$ & $0.06\pm .04$ & $0.00\pm .00$ & $0.11\pm .06$ \\
  \cmidrule(l){2-9}
  & \multirow{5}{*}{\rotatebox[origin=c]{90}{Heuristic}}  & GNN & $3.71\pm .58$ & $1.23\pm .03$ & $0.16\pm .03$ & $1.25\pm .03$ & $0.01\pm .01$ & $1.89\pm .03$ \\
  &           & GNN2 & $2.14\pm .18$ & $2.38\pm .09$ & $0.03\pm .01$ & $2.37\pm .09$ & $0.03\pm .01$ & $2.39\pm .68$ \\
  &           & Grad & $0.96\pm .34$ & $0.04\pm .07$ & $0.10\pm .02$ & $0.03\pm .03$ & $0.01\pm .00$ & $0.64\pm .62$ \\
  &           & Banz & $0.19\pm .04$ & $0.10\pm .01$ & $0.04\pm .02$ & $0.11\pm .02$ & $0.00\pm .00$ & $0.61\pm .66$ \\
  &           & Rnd & $3.44\pm .74$ & $0.06\pm .19$ & $0.16\pm .03$ & $0.02\pm .02$ & $0.01\pm .01$ & $1.22\pm .01$ \\
  \bottomrule
\end{tabular}}

\subsection{one of 2 2trees}
  \resizebox{\columnwidth}{!}{
\begin{tabular}{c|@{~}c@{~}|@{~}c@{~}|rr|rr|rr}
  \toprule
  \multicolumn{2}{c}{} & & \multicolumn{2}{c|}{Addition} & \multicolumn{2}{c|}{Add+RR} & \multicolumn{2}{c}{Add+RR+FT}\\
  \multicolumn{2}{c}{Search} & Rank. & Excess & Time [s] & Excess & Time [s] & Excess & Time [s]\\  
  \midrule
  \parbox[t]{2mm}{\multirow{6}{*}{\rotatebox[origin=c]{90}{Flat}}} 

 & Greedy    & -         & $0.70\pm .21$ & $0.68\pm .20$ & $0.06\pm .03$ & $0.71\pm .21$ & $0.04\pm .02$ & $1.42\pm .41$ \\
  \cmidrule(l){2-9}
  & \multirow{5}{*}{\rotatebox[origin=c]{90}{Heuristic}} & GNN & $21.43\pm .47$ & $1.03\pm .01$ & $0.11\pm .03$ & $1.02\pm .01$ & $0.02\pm .01$ & $1.66\pm .23$ \\
  & 			    & GNN2 & $2.29\pm .18$ & $1.62\pm .06$ & $0.05\pm .02$ & $1.65\pm .06$ & $0.03\pm .01$ & $2.32\pm .26$ \\
  &           & Grad & $3.00\pm .99$ & $0.01\pm .00$ & $0.04\pm .02$ & $0.02\pm .00$ & $0.02\pm .01$ & $0.72\pm .26$ \\
  &           & Banz & $0.26\pm .10$ & $0.06\pm .00$ & $0.04\pm .02$ & $0.06\pm .00$ & $0.02\pm .01$ & $0.71\pm .23$ \\
  &           & Rnd & $23.46\pm .75$ & $0.04\pm .00$ & $0.11\pm .03$ & $0.08\pm .01$ & $0.02\pm .01$ & $0.79\pm .25$ \\
  \midrule
  \parbox[t]{2mm}{\multirow{6}{*}{\rotatebox[origin=c]{90}{Level-by-level}}}  
  & Greedy    & -        & $0.30\pm .19$ & $0.05\pm .07$ & $0.04\pm .02$ & $0.03\pm .02$ & $0.04\pm .02$ & $0.05\pm .02$ \\
  \cmidrule(l){2-9}
  & \multirow{5}{*}{\rotatebox[origin=c]{90}{Heuristic}}  & GNN & $4.74\pm .55$ & $0.99\pm .01$ & $0.21\pm .05$ & $0.98\pm .01$ & $0.02\pm .01$ & $1.56\pm .18$ \\
  &           & GNN2 & $1.71\pm .13$ & $1.58\pm .06$ & $0.06\pm .03$ & $1.67\pm .06$ & $0.05\pm .02$ & $1.66\pm .06$ \\
  &           & Grad & $0.51\pm .15$ & $0.01\pm .00$ & $0.12\pm .03$ & $0.01\pm .00$ & $0.02\pm .01$ & $0.59\pm .19$ \\
  &           & Banz & $0.11\pm .04$ & $0.06\pm .00$ & $0.03\pm .02$ & $0.06\pm .00$ & $0.02\pm .01$ & $0.55\pm .17$ \\
  &           & Rnd & $4.77\pm .50$ & $0.01\pm .00$ & $0.22\pm .05$ & $0.01\pm .00$ & $0.03\pm .01$ & $0.53\pm .17$ \\
  \bottomrule
\end{tabular}}

\section{Experimental problems on task}\label{app:schemas}
\subsection*{Hepatitis}
{\scriptsize 
\begin{Verbatim}[commandchars=\\\{\}]
\textbf{[Dict]} (present 500 times)
  \textcolor{OliveGreen}{bio}: \textbf{[List]} (present 500 times)
    \textbf{[Dict]} (present 621 times)
       \textcolor{OliveGreen}{activity}: String (5 unique out of 621)
       \textcolor{OliveGreen}{fibros}: String (5 unique out of 621)
  \textcolor{OliveGreen}{indis}: \textbf{[List]} (present 496 times)
    \textbf{[Dict]} (present 5691 times)
      \textcolor{OliveGreen}{alb}: Int64 (2 unique out of 5691)
      \textcolor{OliveGreen}{che}: String (10 unique out of 5691)
      \textcolor{OliveGreen}{dbil}: Int64 (2 unique out of 5691)
      \textcolor{OliveGreen}{got}: String (5 unique out of 5691)
      \textcolor{OliveGreen}{gpt}: String (4 unique out of 5691)
      \textcolor{OliveGreen}{tbil}: Int64 (2 unique out of 5691)
      \textcolor{OliveGreen}{tcho}: String (4 unique out of 5691)
      \textcolor{OliveGreen}{tp}: String (4 unique out of 5691)
      \textcolor{OliveGreen}{ttt}: String (6 unique out of 5691)
      \textcolor{OliveGreen}{ztt}: String (6 unique out of 5691)
  \textcolor{OliveGreen}{inf}: \textbf{[List]} (present 196 times)
    \textbf{[Dict]} (present 196 times)
      \textcolor{OliveGreen}{dur}: String (5 unique out of 196)
  \textcolor{OliveGreen}{sex}: Int64 (2 unique out of 500)    
\end{Verbatim}
}

\subsection*{Mutagenesis}
{\scriptsize
\begin{Verbatim}[commandchars=\\\{\}]
\textbf{[Dict]} (present 188 times)
  \textcolor{OliveGreen}{atoms}: \textbf{[List]} (present 188 times)
    \textbf{[Dict]} (present 4893 times)
      \textcolor{OliveGreen}{atom_type}: Int64 (36 unique out of 4893)
      \textcolor{OliveGreen}{bonds}: \textbf{[List]} (present 4893 times)
        \textbf{[Dict]} (present 10486 times)
          \textcolor{OliveGreen}{atom_type}: Int64 (36 unique out of 10486)
          \textcolor{OliveGreen}{bond_type}: Int64 (6 unique out of 10486)
          \textcolor{OliveGreen}{charge}: Float64 (444 unique out of 10486)
          \textcolor{OliveGreen}{element}: String (7 unique out of 10486)
      \textcolor{OliveGreen}{charge}: Float64 (444 unique out of 4893)
      \textcolor{OliveGreen}{element}: String (7 unique out of 4893)
  \textcolor{OliveGreen}{ind1}: Int64 (2 unique out of 188)
  \textcolor{OliveGreen}{inda}: Int64 (2 unique out of 188)
  \textcolor{OliveGreen}{logp}: Float64,Int64 (107 unique out of 188)
  \textcolor{OliveGreen}{lumo}: Float64 (177 unique out of 188)
  \textcolor{OliveGreen}{mutagenic}: Int64 (2 unique out of 188)
\end{Verbatim}
}

\subsection*{DeviceID}
{\scriptsize
\begin{Verbatim}[commandchars=\\\{\}]
\textbf{[Dict]} (present 57906 times)
  \textcolor{OliveGreen}{device_class}: String (13 unique out of 57906)
  \textcolor{OliveGreen}{device_id}: String (10000+ unique out of 57906)
  \textcolor{OliveGreen}{dhcp}: \textbf{[List]} (present 13505 times)
    \textbf{[Dict]} (present 13505 times)
      \textcolor{OliveGreen}{classid}: String (111 unique out of 6071)
      \textcolor{OliveGreen}{paramlist}: String (138 unique out of 13277)
  \textcolor{OliveGreen}{ip}: String (9498 unique out of 56316)
  \textcolor{OliveGreen}{mac}: String (10000+ unique out of 57906)
  \textcolor{OliveGreen}{mdns_services}: \textbf{[List]} (present 30880 times)
    String (359 unique out of 76845)
  \textcolor{OliveGreen}{services}: \textbf{[List]} (present 49793 times)
    \textbf{[Dict]} (present 281555 times)
      \textcolor{OliveGreen}{port}: Int64 (7854 unique out of 281555)
      \textcolor{OliveGreen}{protocol}: String (2 unique out of 281555)
  \textcolor{OliveGreen}{ssdp}: \textbf{[List]} (present 25350 times)
    \textbf{[Dict]} (present 394328 times)
      \textcolor{OliveGreen}{location}: String (10000+ unique out of 337054)
      \textcolor{OliveGreen}{nt}: String (10000+ unique out of 393989)
      \textcolor{OliveGreen}{server}: String (1067 unique out of 330689)
      \textcolor{OliveGreen}{st}: String (10000+ unique out of 56687)
      \textcolor{OliveGreen}{user_agent}: String (474 unique out of 12016)
  \textcolor{OliveGreen}{upnp}: \textbf{[List]} (present 22881 times)
    \textbf{[Dict]} (present 35616 times)
      \textcolor{OliveGreen}{device_type}: String (96 unique out of 35616)
      \textcolor{OliveGreen}{manufacturer}: String (183 unique out of 35261)
      \textcolor{OliveGreen}{model_description}: String (706 unique out of 22382)
      \textcolor{OliveGreen}{model_name}: String (2913 unique out of 35359)
      \textcolor{OliveGreen}{services}: \textbf{[List]} (present 32712 times)
        String (141 unique out of 78924)
\end{Verbatim}
}

\section{Grid-search for hyper-parameters of GNN}
The table below shows average number of excessive subtrees in the explanation. The numbers are averages over all 3 datasets, seven tasks, and 20 variation of GNN explainer with HAdd subtree selection method. $\alpha$ is the penalization of entropy and $\beta$ is the penalization of sum of the mask.

\begin{center}
\begin{tabular}{lrrr}
& $\alpha$ & $\beta$  &\textbf{excess leaves} \\
\hline
gnn & 1 & 0.1 & $4.07\pm .12$ \\
gnn & 1 & 0.005 & $36.77\pm .69$ \\
gnn & 0.1 & 0.005 & $6.52\pm .17$ \\
gnn & 0.01 & 0.005 & $9.14\pm .54$ \\
gnn & 0.1 & 0.1 & $15.61\pm .67$ \\
gnn & 0.01 & 0.1 & $4.79\pm .33$ \\
gnn & 1 & 0.05 & $6.40\pm .17$ \\
gnn & 0.1 & 0.05 & $15.31\pm .66$ \\
gnn & 0.01 & 0.05 & $5.30\pm .36$ \\
gnn & 1 & 0.01 & $22.97\pm .43$ \\
gnn & 0.1 & 0.01 & $4.18\pm .11$ \\
gnn & 0.01 & 0.01 & $7.30\pm .47$ \\
gnn & 0.1 & 1 & $12.23\pm .52$ \\
gnn & 0.01 & 1 & $7.16\pm .36$ \\
gnn & 1 & 0.5 & $29.71\pm .84$ \\
gnn & 0.1 & 0.5 & $13.85\pm .59$ \\
gnn & 0.01 & 0.5 & $5.89\pm .33$ \\ \hline \hline
\end{tabular}
\end{center}

\section{Compute hardware}\label{app:hw}
All the experiments were performed on a uniform cluster of nodes with 56 \texttt{Intel(R) Xeon(R) Gold 5120 CPU @ 2.20GHz}, \texttt{376 GB} of memory per node, running \texttt{CentOS Linux 7}. The jobs were scheduled through \texttt{slurm 19.05.5}, limiting each job to a single CPU and 20GB of memory.

\onecolumn
\section{Examples of explanation in Cuckoo dataset}\label{app:cuckoo}
Two figures below show explanations of two clean samples from Cuckoo dataset by the proposed the proposed Level-By-Level selection with random removal and Banzhaf score and by GNN explainer. The explanation by the proposed approach are consistent and provide the user with a clear indication that his classifier has overfit to the time of sample's execution in the sandbox. Thanks to this, user can remove these keys and retrain his classifier (and repeat the investigation for furher bias in the dataset).

\begin{figure}[bh!]
\begin{subfigure}{\columnwidth}
\begin{minted}[fontsize=\tiny]{json}
{"info": {
    "machine": {
      "started_on": "2019-07-02 23:09:00"}}}
\end{minted}
\caption{LbyL-Banz-Add+RR}
\end{subfigure}\\
\begin{subfigure}{\columnwidth}
\begin{minted}[fontsize=\tiny]{json}
{ "network": {
    "domains": [
      {
        "domain": {
          "and": [
            "dns.msftncsi.com",
            "teredo.ipv6.microsoft.com"
          ]
        },
        "ip": "131.107.255.255"
      }
    ],
    "dns": [
      {
        "type": "A",
        "request": {
          "and": [
            "teredo.ipv6.microsoft.com",
            "time.windows.com"
          ]
        }
      }
    ],
    "pcap_sha256": "2dd41ebbaeb590e4164a918d6f8eb5406d19d0e45aaf104565c73736630a6fe6"
  },
  "static": {
    "imported_dll_count": 0
  },
  "signatures": [
    {
      "markcount": 1
    }
  ],
  "target": {
    "file": {
      "crc32": "0BBED2CB",
      "sha256": "50fc5048e9c523ad0ad01ee5733a05195b9d14414e13395f86407259e9a923c4"
    },
    "category": "file"
  },
  "info": {
    "git": {
      "fetch_head": "13cbe0d9e457be3673304533043e992ead1ea9b2"
    },
    "machine": {
      "shutdown_on": "2019-07-02 23:12:21"
    },
    "options": "procmemdump=yes",
    "platform": "windows",
    "monitor": "2deb9ccd75d5a7a3fe05b2625b03a8639d6ee36b",
    "category": "file"}}
\end{minted}
\caption{GNN explainer}
\end{subfigure}
\caption{Explanation of one clean sample from Cuckoo dataset by (a) the proposed Level-By-Level selection with random removal and Banzhaf score and (b) by GNN explainer.}
\end{figure}

\begin{figure}[h!]
\begin{subfigure}{\columnwidth}
\begin{minted}[fontsize=\tiny]{json}
{
  "info": {
    "machine": {
      "shutdown_on": "2019-07-02 23:29:54"
    }
  }
}
\end{minted}
\caption{LbyL-Banz-Add+RR}
\end{subfigure}\\
\begin{subfigure}{\columnwidth}
\begin{minted}[fontsize=\tiny,breaklines]{json}
{
  "metadata": {
    "output": {
      "pcap": {
        "basename": "dump.pcap",
        "sha256": "da6dac4c795ca4fe32f9b04e4e61cdc55633cc032f9306f0b55d3ddeea63f82b"
      }
    }
  },
  "network": {
    "domains": [
      {
        "domain": {
          "and": [
            "dns.msftncsi.com",
            "teredo.ipv6.microsoft.com"
          ]
        },
        "ip": "131.107.255.255"
      }
    ],
    "dns": [
      {
        "type": "A"
      }
    ]
  },
  "signatures": [
    {
      "ttp": {
        "T1045": {
          "long": "Software packing is a method of compressing or encrypting an executable. Packing an executable changes the file signature in an attempt to avoid signature-based detection. Most decompression techniques decompress the executable code in memory.",
          "short": "Software Packing"
        }
      },
      "markcount": {
        "and": [
          1,
          4
        ]
      },
      "description": "The file contains an unknown PE resource name possibly indicative of a packer",
      "name": "origin_langid",
      "severity": {
        "and": [
          1,
          2
        ]
      }
    }
  ],
  "target": {
    "file": {
      "sha1": "c5c35d002f72a464e19aa01b23b334d9a67e3fc3",
      "scalars": {
        "1": "[0.0005075704]"
      },
      "name": "filename",
      "sha256": "d999e413ab353b5127ed7c3f889e8b223e64d7786465a048ae6d8c3df80d4fee",
      "sha512": "88316d4d84dbcd321e2678c6cf576231abc2ee28a646bf7b3faaa3c266cbb0bf5e6e352476925188c630cdb35e87fb2e521292e289e52298edca9cfc39e67867"
    },
    "category": "file"
  },
  "info": {
    "machine": {
      "label": "Windows7",
      "shutdown_on": "2019-07-02 23:29:54"
    },
    "score": 0.8,
    "category": "file"
  }
}
\end{minted}
\caption{GNN explainer}
\end{subfigure}
\caption{Explanation of one clean sample from Cuckoo dataset by (a) the proposed Level-By-Level selection with random removal and Banzhaf score and (b) by GNN explainer.}

\end{figure}

\section{Reproducibility checklist}\label{app:checklist}
\begin{itemize}
    \item\textbf{Datasets} All artificial datasets contained 10 000 samples and they are provided in the supplementary material. Since the goal of the paper is explanation, we have used all samples for training and selected randomly 100 for explanation. The seed for random number generator was set to the index of variant of the dataset (for each dataset and task, there are 20 variants). For the explanation, we have used samples that were correctly classified by the classifier. We provide only a link to the datasets in Appendix~\ref{app:zip} and not the full datasets because of their size.
    
    \item\textbf{Libraries} The code dependencies are automatically handled by Julia package manager. To restore the experimental environment, we recommend to use Julia 1.4.1 and execture 
    \texttt{julia --profile -e "using Pkg;Pkg.instantiate"} in directory \texttt{ExplainMill.jl/scripts}. This restores the exact environments (up-to git-commits).
    \item\textbf{Commands} The commands to recreate results are provided in files \\ \texttt{ExplainMill.jl/scripts/datasets/ artificial/submit.sh} and \\ \texttt{ExplainMill.jl/scripts/cuckoo/submit.sh}, which issues approriate jobs on a slurm clusters.
    
    \item\textbf{Tables}  Code to recreate tables is in \texttt{ExplainMill.jl/scripts/datasets/tables.jl}.
    
    \item\textbf{Hyper-parameters} Most of the parameters are reported in the paper of this appendix. However, some details are present only in the source codes. Examples of these include the parameters of the ADAM optimizer for GNN explainer by paper, which we reused from their published source codes. Specific random seeds are also present only in the supplied zip file, since they would not be meaningful without the specific random numbers generator used. We also do not report details on the search for the used number of samples for estimating Banzhaf values, since the reported experiments clearly show that the used heuristic has small influence on the overall performance, compared to selection of the right search technique. We hand-picked this value based on initial experiments only.
\end{itemize}



\end{document}